\documentclass[10pt,journal,compsoc]{IEEEtran}
\def\IEEEcompsocdiamondline{}

\usepackage{ifpdf}
\ifpdf
  \pdfoutput=1\relax                   
\else
\fi%

\usepackage[numbers,sort&compress]{natbib}

\ifCLASSINFOpdf
  \usepackage[pdftex]{graphicx}
  \DeclareGraphicsExtensions{.pdf,.jpeg,.png}
\else
  \usepackage[dvips]{graphicx}
  \DeclareGraphicsExtensions{.eps}
\fi

\graphicspath{{figures/}{pictures/}{images/}{./}} 
\usepackage{comment}
\usepackage{amsmath,amssymb} 
\usepackage{color}
\usepackage{capt-of}
\usepackage{cuted}
\usepackage{xcolor}
\usepackage{hyperref}
\usepackage{subcaption}
\usepackage{wrapfig}
\usepackage{tabu}
\usepackage{tcolorbox}
\usepackage{scalerel}
\usepackage{xspace,xpunctuate}

\newcommand{\ie}{{i.e.,}\xspace}
\newcommand{\eg}{{e.g.,}\xspace}
\newcommand{\cf}{{cf.}\xspace}
\newcommand{\ea}{{et~al\xperiod}\xspace}

\definecolor{figurecolor}{HTML}{9E9E9E}
\newtcbox{\figurebox}{on line, colframe=figurecolor,colback=figurecolor!10!white, boxrule=0.5pt,arc=4pt,boxsep=0pt,left=2pt,right=2pt,top=2pt,bottom=2pt,fontupper=\sffamily} 
\definecolor{objectivecolor}{HTML}{2196F3}
\newtcbox{\objectivebox}{on line, colframe=objectivecolor,colback=objectivecolor!10!white, boxrule=0.5pt,arc=4pt,boxsep=0pt,left=2pt,right=2pt,top=2pt,bottom=2pt}
\definecolor{technicalcolor}{HTML}{f44336}
\newtcbox{\technicalbox}{on line, colframe=technicalcolor,colback=technicalcolor!10!white, boxrule=0.5pt,arc=4pt,boxsep=0pt,left=2pt,right=2pt,top=2pt,bottom=2pt}
\definecolor{visualizationcolor}{HTML}{FF9800}
\newtcbox{\visualizationbox}{on line, colframe=visualizationcolor,colback=visualizationcolor!10!white, boxrule=0.5pt,arc=4pt,boxsep=0pt,left=2pt,right=2pt,top=2pt,bottom=2pt}
\definecolor{guidecolor}{HTML}{8BC34A}
\newtcbox{\guidebox}{on line, colframe=guidecolor,colback=guidecolor!10!white, boxrule=0.5pt,arc=4pt,boxsep=0pt,left=2pt,right=2pt,top=2pt,bottom=2pt} 
\definecolor{hypothesiscolor}{HTML}{8BC34A}
\newtcbox{\hypothesisbox}{on line, colframe=hypothesiscolor,colback=hypothesiscolor!10!white, boxrule=0.5pt,arc=4pt,boxsep=0pt,left=2pt,right=2pt,top=2pt,bottom=2pt} 
\definecolor{loadcolor}{HTML}{9c27b0}
\newtcbox{\loadbox}{on line, colframe=loadcolor,colback=loadcolor!10!white, boxrule=0.5pt,arc=4pt,boxsep=0pt,left=2pt,right=2pt,top=2pt,bottom=2pt} 

\newcommand{\enn}{\textit{exploRNN}}

\begin{document}
\title{
  exploRNN:\\
  Teaching Recurrent Neural Networks\\
  through Visual Exploration
}

%
\author{Alex Bäuerle,
        Patrick Albus,
        Raphael Störk,
        Tina Seufert,
        and Timo Ropinski
\IEEEcompsocitemizethanks{\IEEEcompsocthanksitem All authors were with Ulm University, Germany\protect\\
Email: see \url{https://a13x.io}.}
}

%



\IEEEteaser{
    \centering
    \includegraphics[width=0.9\linewidth]{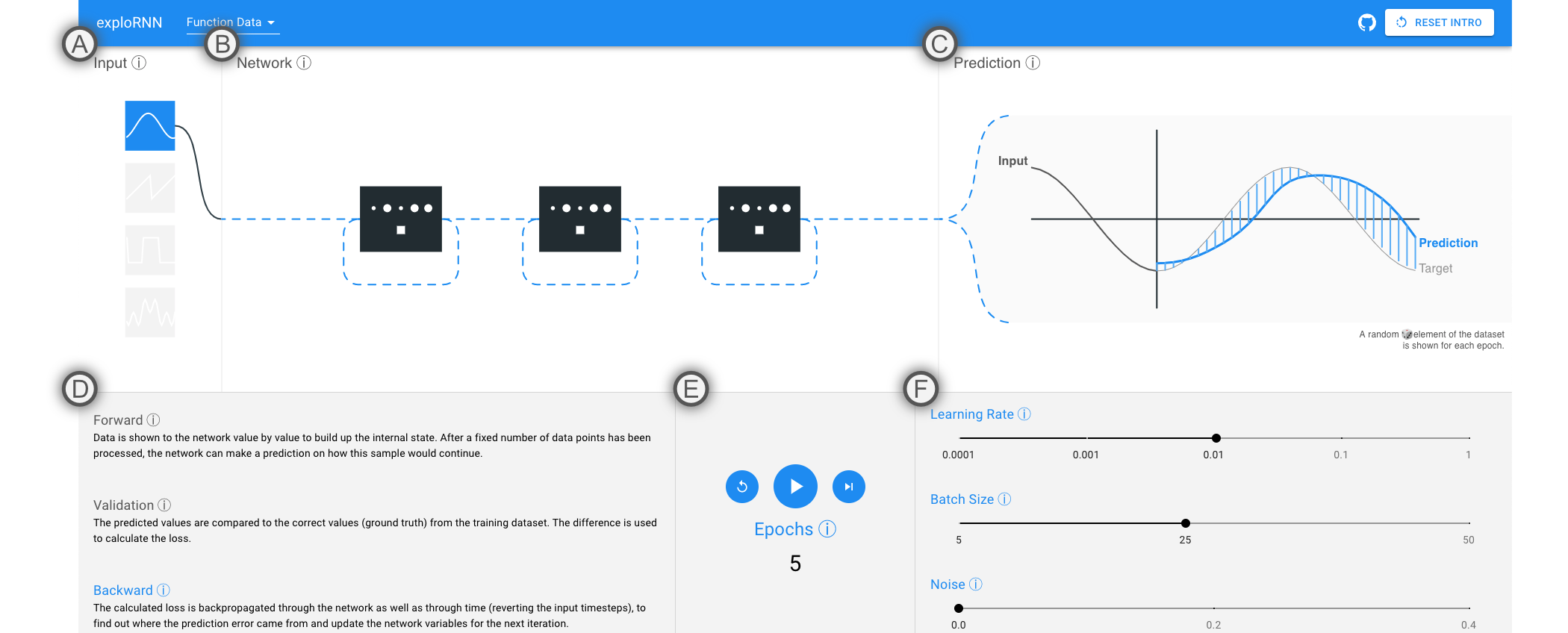}
    \setcounter{figure}{0}
    \captionof{figure}{
    \figurebox{A} Simple input types illustrate the abstract concepts behind RNNs.
    \figurebox{B} An animated, modifiable network architecture shows the data flow.
    \figurebox{C} The prediction visualization shows the network input, prediction, ground truth, and error bars, all animated to communicate their temporal nature.
    \figurebox{D} Text helps explain the training process.
    \figurebox{E} RNNs can be interactively trained.
    \figurebox{F} Training parameters can be interactively explored.
    \label{fig:networkview} 
    }
}

\IEEEtitleabstractindextext{%
\begin{abstract}
Due to the success and growing job market of deep learning (DL), students and researchers from many areas are interested in learning about DL technologies.
Visualization has been used as a modern medium during this learning process.
However, despite the fact that sequential data tasks, such as text and function analysis, are at the forefront of DL research, there does not yet exist an educational visualization that covers recurrent neural networks (RNNs).
Additionally, the benefits and trade-offs between using visualization environments and conventional learning material for DL have not yet been evaluated.
To address these gaps, we propose~\enn, the first interactively explorable educational visualization for RNNs.
\enn is accessible online and provides an overview of the training process of RNNs at a coarse level, as well as detailed tools for the inspection of data flow within LSTM cells.
In an empirical between-subjects study with 37 participants, we investigate the learning outcomes and cognitive load of \enn~compared to a classic text-based learning environment.
While learners in the text group are ahead in superficial knowledge acquisition, \enn~is particularly helpful for deeper understanding.
Additionally, learning with \enn~is perceived as significantly easier and causes less extraneous load.
In conclusion, for difficult learning material, such as neural networks that require deep understanding, interactive visualizations such as \enn~can be helpful.
\end{abstract}

}

\maketitle

\IEEEdisplaynontitleabstractindextext
\IEEEpeerreviewmaketitle

\section{Introduction}\label{sec:introduction}

With its recent advances, DL has gained immense traction in research, industry, and education.
As job opportunities related to machine learning are unprecedented, many want to learn about and understand DL technologies.

While initial progress in DL was mainly possible due to the rise of convolutional neural networks (CNNs), large training data sets, and GPU training in the context of image recognition~\cite{krizhevsky2012imagenet,he2016deep,szegedy2015going}, other network architectures, such as RNNs, which are able to process sequential data, are becoming increasingly important.
At the same time, these more advanced learning architectures are more difficult to comprehend, as they employ concepts that are fundamentally different from classical computer science. 
Thus, by making the process behind RNNs transparent and easy to understand, research in sequential learning tasks can be accelerated as the field opens up to additional users and contributors.

Along this line, the visualization community has shown how interactive visual explorables can be effective for learning about DL concepts~\cite{kahng2018gan,norton2017adversarial,convnetjs,smilkov2017direct}.
Since different architectures come with their unique challenges, existing educational applications usually focus on one type of architecture.
Unfortunately, the set of existing applications still does not cover RNNs.
This is despite the fact that RNNs are widely adopted in tasks such as speech processing~\cite{graves2013speech,sak2014long}, handwriting recognition~\cite{graves2008novel}, and machine translation~\cite{wu2016google}, among many others.
While RNNs are capable of solving such sequential tasks, they also bring their unique architectures and concepts to capture temporal information.
As these concepts differ from other network types, RNN education could be of great benefit.
To facilitate RNN education, we propose \enn, an interactive explorable visualization for RNNs that runs directly in any modern web browser.

The focus of \enn~is to make learning about these abstract and complex network types easier, more motivating, and more applicable to real problems.
By presenting learning material in a way that is conducive to learning, learners should need fewer unnecessary cognitive resources \loadbox{CR}~\cite{sweller2011cognitive}.
These freed-up resources are then available to be used for a deeper understanding \loadbox{DU} of the learning material.
We also expect that this would result in a more motivating and joyful \loadbox{MJ} learning experience compared to traditional learning methods, such as text.
In turn, learners might be willing to spend more time learning, and more learners could be attracted in general, effectively increasing overall knowledge gain.
To assess these hypotheses, we compare \enn{} with text-based learning in a between-subjects study with 37 participants.
Our evaluation provides insights into when, and under which conditions, visual interactive learning environments can outperform conventional learning material.

Along this line, we make the following contributions:

\begin{itemize}
  \item \textbf{Educational Objectives and Design Challenges} for educational RNN visualizations, informing our visualization design.
  \item \textbf{An interactive visualization approach for RNN education}, enabling investigation at different levels of granularity.
  \item \textbf{A quantitative, comparative evaluation}, investigating the effectiveness of our approach and providing hints for other interactive educational visualizations.
\end{itemize}

\enn~can be accessed online at:~\url{https://mi-pages.informatik.uni-ulm.de/explornn}, contributing to a fast-growing corpus of visualization work in the field of DL.
To our knowledge, \enn~is the first educational visualization interface that is targeted at RNNs, an important and growing class of neural networks (NN).
Additionally, our study is the first to compare conventional learning material to a visual, interactive learning environment for DL education.

\section{Background: RNNs}\label{sec:background}

We would like to invite readers who want to refresh their knowledge on RNNs to use \enn~at~\url{https://mi-pages.informatik.uni-ulm.de/explornn} as an interactive learning experiment.
This chapter contains a brief summary of the knowledge that is communicated in~\enn.

CNNs and multi-layer perceptrons (MLPs), which are used for most classical DL tasks, process data in a feedforward manner.
On the contrary, RNNs provide a cyclical architecture in which the output of the previous timestep is used in combination with new inputs to inform the activation of a cell.
The main difference in training RNNs is backpropagation through time (BPTT), where the prediction error is propagated not only back through the layers but also within the recurrent connections of the layers.

\begin{figure}[t]
  \centering
  \includegraphics[width=0.6\linewidth]{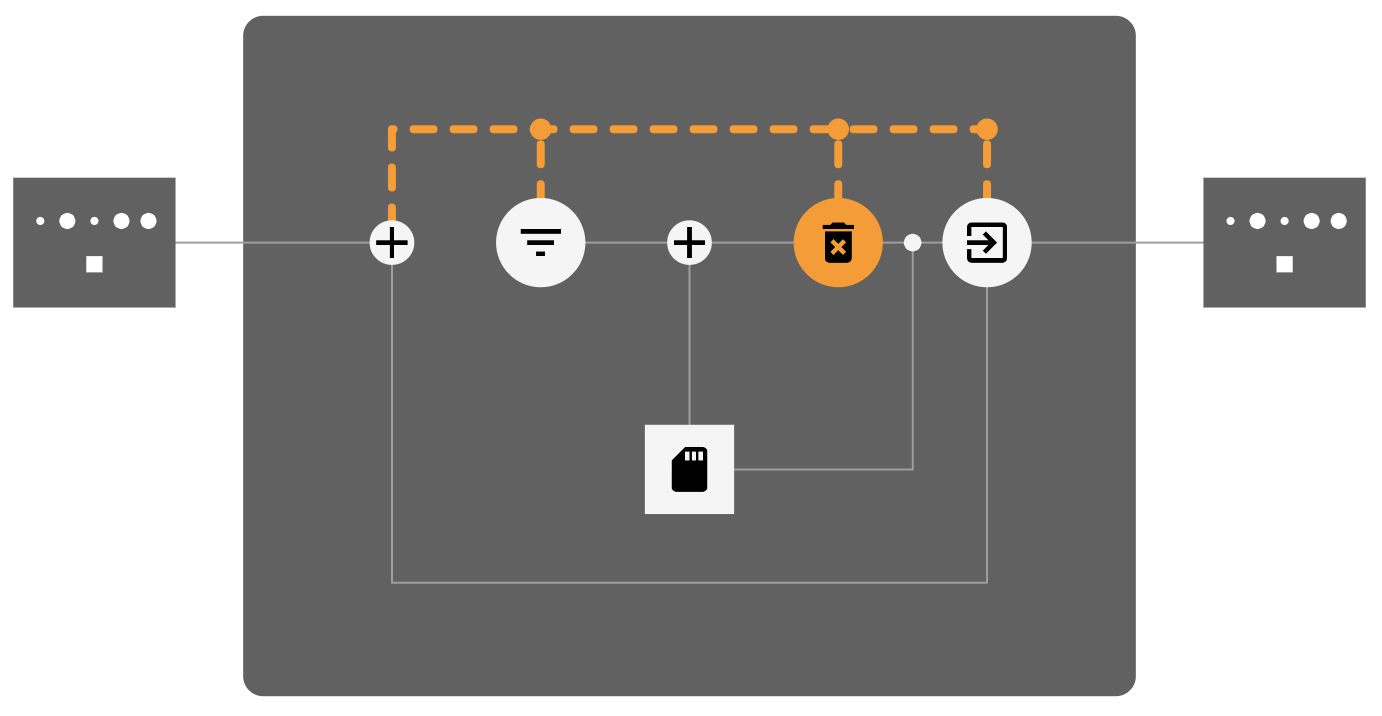}
  \caption{
  LSTM cell with all operations visualized.
  The input is added to the output of the previous time step and then used by the three gates for the gate activation.
  }
  \label{fig:lstm}
\end{figure}

We visualize the LSTM architecture (\cf{} \autoref{fig:lstm}).
Although this is not the most simple recurrent architecture that exists, it is superior in capturing long-range dependencies, as it mitigates the vanishing gradient problem~\cite{hochreiter1997long,bengio1994learning,pascanu2013difficulty}, and is thus widely used.
The main features of an LSTM cell are the gating mechanisms and the cell state.
The three gates within an LSTM cell are computed based on the input at time step $t$, $x^t$ and the activation of the cell at time step $t-1$, $a^{t-1}$ as follows:
\\
\noindent\textbf{\scalerel*{\includegraphics{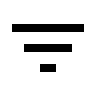}}{)} Input Gate:}
$i^t = sigmoid(W_{ix}x^t + W_{ia}a^{t-1} + b_i)$, what new information to use to update the cell state $c^t$.
\\
\noindent\textbf{\scalerel*{\includegraphics{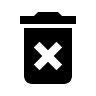}}{)} Forget Gate:}
$f^t = sigmoid(W_{fx}x^t + W_{fa}a^{t-1} + b_f)$, what information in the cell state $c^t$ can be forgotten.
\\
\noindent\textbf{\scalerel*{\includegraphics{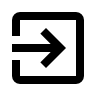}}{)} Output Gate:}
$o^t = sigmoid(W_{ox}x^t + W_{oa}a^{t-1} + b_o)$, what part of the cell state $c^t$ is used to compute the activation.
\\
The cell state \scalerel*{\includegraphics{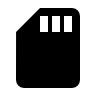}}{)} at timestep $t$ is then computed as $c^t = f^t \circ c^{t-1} + i^t \circ tanh(W_{cx}x^t + W_{ca}a^{t-1} + b_c)$, where $\circ$ is the hadamard product.

While there are other architectures that also use the concept of cell state and modular updating, such as gated recurrent units (GRUs)~\cite{cho2014learning}, their underlying idea does not greatly differ.
However, since LSTMs were the first to introduce the explained concepts, are more general in their application~\cite{weiss2018practical}, and often outperform GRUs~\cite{britz2017massive}, we focus on conveying the LSTM architecture.

\section{Related Work}
In this section, we first give a brief overview of the explorable explanation literature before elaborating on the corpus of related work in the area of educational visualizations for DL non-experts and RNN visualizations.

\subsection{Explorable Explanations}
Explorable learning environments were invented long before DL raised awareness in the broader public.
Their effectiveness was investigated in the line of work by Hundhausen \ea{}~\cite{hundhausen2007you,hundhausen2002meta}.
Schweitzer and Brown then described design characteristics and an evaluation of active learning settings in classrooms by using visualization~\cite{schweitzer2007interactive}.
There also exists a line of work on the use of visualization for programming education~\cite{guo2013online,guo2015codechella,drosos2020wrex}.
These approaches show how visualization can communicate algorithmic thinking effectively.
We combine these ideas with more recent concepts, which have been proposed under the term \textit{exploranation} in the area of science education~\cite{ynnerman2018exploranation}, where explorable explanations provide benefits for learning.

There are numerous helpful visualizations conveying properties of NN architectures, their functionality, or application scenarios~\cite{harley2015interactive,convnetjs,norton2017adversarial}.
However, we will focus on educational, explorable visualization approaches that have been proposed for different network types.
One of the most prominent interactive educational visualization approaches has been proposed by Smilkov \ea{}~\cite{smilkov2017direct}.
In their explainable \textit{Tensorflow Playground}, one can select the properties of a NN to be trained.
They also allow the customization of certain training parameters and deployed their approach as a web-based application.
Similarly, in \textit{Revacnn}, users can explore the activations of a CNN by modifying the network structure and training the network in the browser~\cite{chung2016revacnn}.
While these approaches help teach the most basic concepts of MLPs and CNNs, respectively, more advanced architectures need further, specialized visualizations.
Another approach that is closely related to ours, but works on a different type of NNs, namely, GANs, is \textit{GanLab}~\cite{kahng2018gan}.
They focus on how the generator and discriminator are used adversarially to yield synthetic data that resembles the data distribution it was trained on.
However, GANs bring their own visualization challenges, which are fundamentally different from those we found for RNNs.
Additionally, neither of these systems was systematically and quantitatively evaluated.

As none of these visualization approaches is designed to help non-experts understand how RNNs function, with their unique concepts of memory and temporal dependence, our aim is to fill this gap in the literature with \enn.
Additionally, we shed light on the usefulness of such interactive learning environments through our quantitative user study.
We specifically examine the difference in learning outcomes across different learning hierarchies~\cite{bloom1956taxonomy}, the complexity of the learning experience by means of cognitive load~\cite{sweller2011cognitive}, and qualitative assessments such as motivation, perceived quality of content, and joy throughout the learning process.
We hope that our findings in this area can be an indication for similar learning environments and motivate others to conduct similar experiments.

\subsection{RNN Visualizations}
Apart from educational visualizations, there is another line of visualization work targeted towards investigating RNNs.
These approaches are designed mainly for researchers who want to understand and debug their models.
An early approach towards visualizing RNNs was proposed by Karpathy \ea{}, who visualize the activation of RNN cells for expert analysis~\cite{karpathy2015visualizing}.
Strobelt \ea{} published \textit{LSTMVis}, in which the hidden state dynamics of RNNs is investigated~\cite{strobelt2017lstmvis}.
They specifically demonstrate how text understanding can be analyzed through investigating the structure and change of the cell state.
They also presented \textit{Seq2Seq-Vis}, in which sequence-to-sequence models can be probed to reveal errors and learned patterns~\cite{strobelt2018s}.
Along the same line, Ming \ea{} introduced RNNVis~\cite{ming2017understanding}.
They analyze the functionality of individual hidden state units by observing their reaction to specific text segments.
With RNNbow, Cashman \ea{} published a visualization, in which the gradients of RNNs can be analyzed~\cite{cashman2018rnnbow}.
They attribute the gradient to individual letters in a textual input sequence.
This way, researchers can inspect how their models learn.
In another approach, Shen \ea{} proposed visualizations for RNNs~\cite{shen2020visual} operating on multi-dimensional sequence data.
Here, developers can inspect hidden unit responses to get insight into different networks.
Similarly, Garcia and Weiskopf proposed a visualization for the inspection of hidden states of RNNs~\cite{garcia2020inner}.
However, all approaches described here are expert tools that help during development.
Contrary to this, we aim to convey the general idea of RNNs to novices in this area of DL.
\\
\\
Insights on the effects of using exploration and visualization for learning in general, as well as present educational visualizations for NNs, show how interactive exploration can help a broader audience with access to learning experiences.
Therefore, we propose \enn, which provides insight into the function of RNNs for users who know the basics of DL, but are laymen in the area of sequential learning.
Our evaluation also provides the first comparative analysis of interactive learning environments and classical learning approaches for NN education.

\section{Educational Objectives}\label{sec:eduobj}

To inform the visualization design of a learning experience, educational objectives are needed, which we defined based on Bloom's taxonomy~\cite{bloom1956taxonomy}.
Our target users already understand the fundamental concepts of DL and know about feed-forward NNs.
Without this background knowledge, the theory behind those techniques would first need to be explained, which would extend the scope of \enn{}.
As our target audience aims to learn the yet unknown concepts of RNNs, we focused on \textit{recall} \objectivebox{O1\&2}, \textit{comprehension} \objectivebox{O2\&3}, and \textit{transfer} \objectivebox{O3\&4} of the learned information.
Later, this learning can be applied in the wild to access levels four to six (\textit{analyze}, \textit{evaluate}, \textit{create}) of Bloom's taxonomy.
Formulated on this basis, our educational objectives are:

\noindent\textbf{\objectivebox{O1} Justification.}
Users should know that RNNs, in contrast to other network types, can be used for sequential data.
This also includes BPTT, through which RNNs can learn temporal dependencies, which classical feed-forward networks cannot.
\\
\noindent\textbf{\objectivebox{O2} Cell Structure.}
Users should then understand how LSTM cells are built and what functionality their individual components have.
Here, the cell gates, as well as the memory element, are of special importance, as they enable the processing of sequential data.
\\
\noindent\textbf{\objectivebox{O3} Training Setup.}
To understand the training process of such networks, users should know important parameters for the setup of RNNs.
This includes the network structure, training parameters, and how data is fed to the network.
\\
\noindent\textbf{\objectivebox{O4} Task.}
Finally, to transfer this theoretical knowledge about RNNs to real applications, users should learn about different application areas and data types that can be used with RNNs.
In the end, they should be able to describe how RNNs could be used for their own application scenarios.

Similarly to a lecture at a university or a textbook, our learning environment is designed to provide an introduction to RNNs from which interested users can start experimenting with the techniques.
Accordingly, our educational objectives not only motivate the importance of RNNs but are also aimed at providing insights about the input data and related tasks, as well as how the training process and LSTM cells work.

\section{Design Challenges}
Since RNN cells are a special form of NN layers, they open up unique challenges for visualization-based education.
We observed both visualization design challenges and technical challenges, which we describe in this section.

\subsection{Visualization Design Challenges}\label{sec:vischallenges}
We first discuss the following visualization design challenges that we identified in the context of an interactive learning environment aimed at RNNs and illustrate how they relate to our educational objectives:

\noindent\textbf{\visualizationbox{V1} Complexity.}
As mentioned in~\autoref{sec:introduction}, one of our central goals is to simplify learning by reducing cognitive load~\cite{sweller2011cognitive}.
However, RNNs are typically trained on a large amount of complex data that can be difficult to grasp \objectivebox{O1} \objectivebox{O4}~\cite{soltau2016neural,manh2018scene}.
The same holds for network architectures, which are also often too complex to fully comprehend in their entirety \objectivebox{O3}~\cite{graves2013hybrid,park2018sequence}.
Consequently, all visualizations must be interpretable and intuitive, but realistic enough to form a compelling use case~\cite{gleicher2013explainers,munzner2014visualization}.
\\
\noindent\textbf{\visualizationbox{V2} Dynamics.}
An educational system to teach RNN concepts should clarify the dynamics of the sequential data on which these networks operate \objectivebox{O1}, as well as the dynamics of the training process \objectivebox{O3}.
These dynamic processes must be visually communicated, including data type and data processing, both forward (inference) and backward (backpropagation), within the network~\cite{beck2014state}.
\\
\noindent\textbf{\visualizationbox{V3} Multiscale.}
RNN structures need to be communicated at different granularities, \ie network, cell, and cell components \objectivebox{O2}.
These multiple scales need to be fluidly inspectable, while at the same time, the granularity at which the user currently operates must be communicated~\cite{shneiderman1996eyes,munzner2014visualization}.
\\
\noindent\textbf{\visualizationbox{V4} Supervision.}
In classical learning settings, teacher supervision or other opportunities to seek further information is provided.
Contrary to this, \enn~is designed as a standalone learning environment that does not require external guidance~\objectivebox{O1-4}.
Thus, supervision has to be substituted by visual guidance~\cite{amadieu2011attention,de2010attention}.

\subsection{Technical Challenges}\label{sec:techchallenges}
Whereas the visualization design challenges are based directly on our educational objectives, the following technical challenges relate to the development of such an interactive, explorable learning environment:

\noindent\textbf{\technicalbox{T1} Training Time.}
Typically, training processes can take up to several days to convergence~\cite{khomenko2016accelerating,zhu2018benchmarking}.
However, for an interactive learning experience, waiting days for convergence is not feasible.
To provide direct feedback to the user, our networks thus have to converge in minutes instead of hours or days.
\\
\noindent\textbf{\technicalbox{T2} Training Steps.}
Normally, computation is done as fast as possible to minimize the time it takes for the network to converge.
However, we want the user to be able to follow the training process and observe individual training steps~\cite{amadieu2011attention}.
Thus, training steps should be separated temporally from the visualization.
\\
\noindent\textbf{\technicalbox{T3} Deployment.}
Modern-day learning is often conducted via online courses, blog posts, or explainable web pages~\cite{peters2000digital}.
Although this makes such learning environments accessible to a broad audience, it also limits the technical freedom of such applications~\cite{smilkov2019tensorflow}.
Therefore, educational environments should be deployed to a broad audience, while also providing diverse functionality.

\section{Visualization Design}\label{sec:interaction}

\begin{figure}[t]
  \centering
  \includegraphics[width=0.6\linewidth]{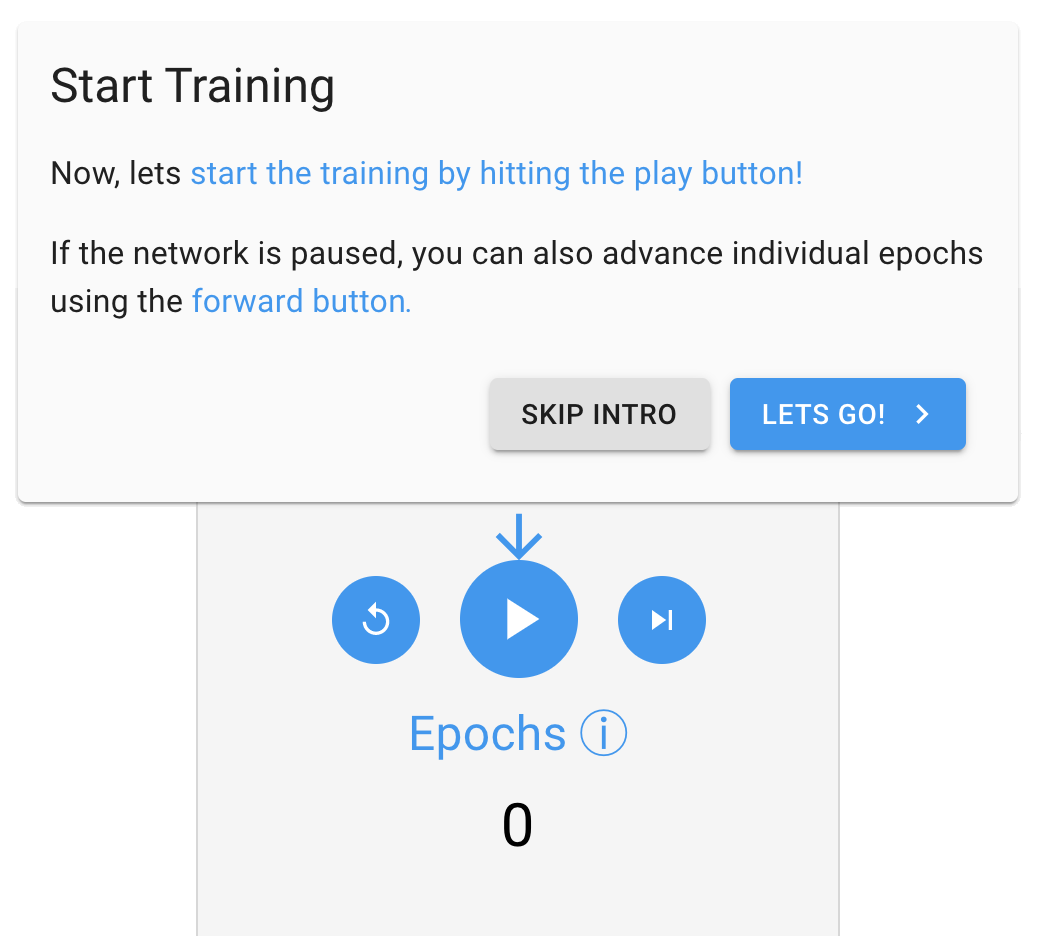}
  \caption{
  Onboarding dialogs guide the user through our visualizations, so that no manual introduction is needed, and the user can explore \enn~on their own.
  Textual descriptions with highlights provide detailed explanations for individual components.
  Positioning and arrows reveal associations between dialogs and components.
  }
  \label{fig:onboarding}
\end{figure}

\begin{figure}[b]
  \centering
  \includegraphics[width=0.8\linewidth]{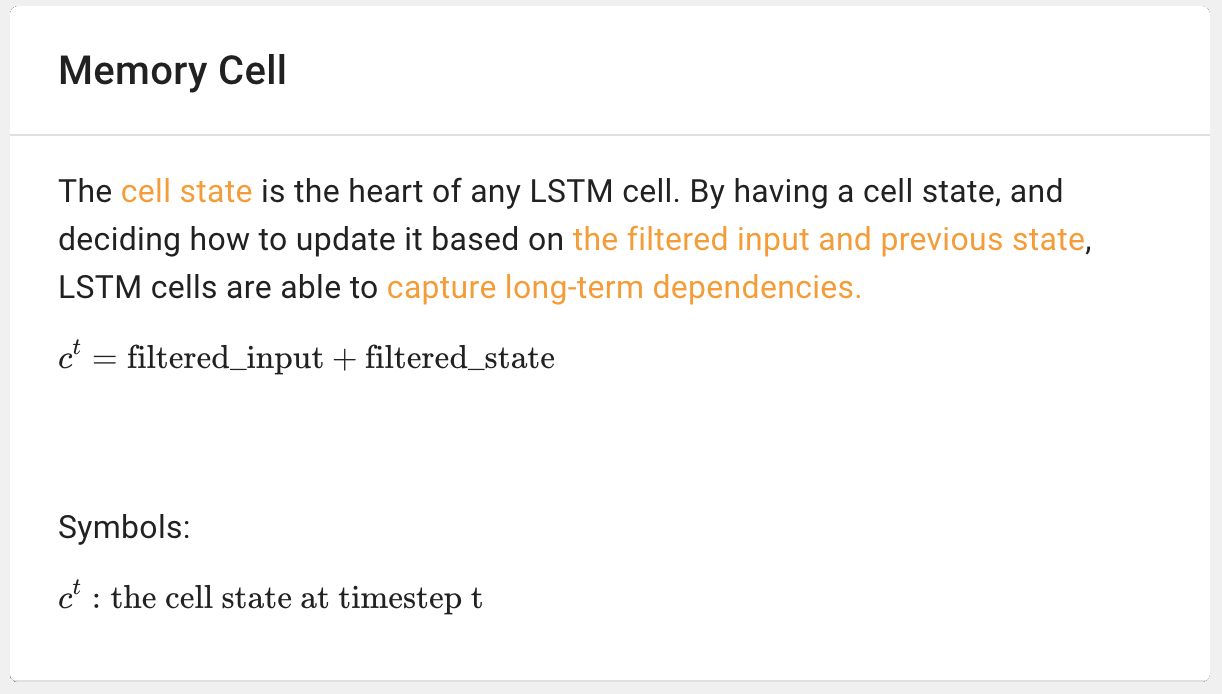}
  \caption{
  Users can access more detailed explanations for many elements of our visualizations such as training steps, hyper-parameters, and operations in a cell.
  }
  \label{fig:explanation}
\end{figure}

\begin{figure*}[t]
  \centering
  \includegraphics[width=\linewidth]{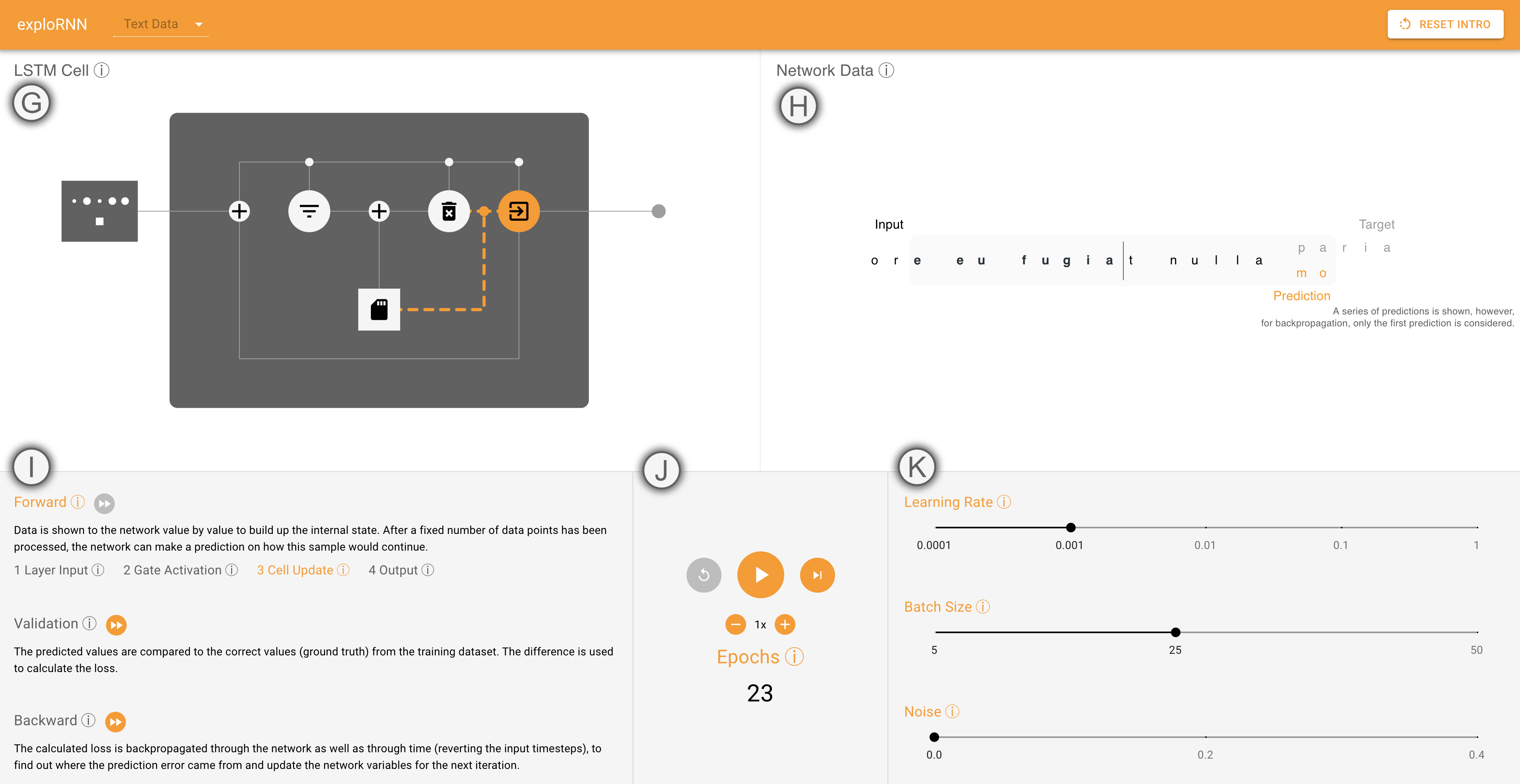}
  \caption{
  LSTM cell view.
  \figurebox{G} Visualization of data flow through the cell.
  \figurebox{H} Input to the network and its prediction. Visualization of the training error computation. A grey sliding window indicates which data points are needed to initialize the cell state.
  \figurebox{I} Explanations with more detailed steps for the forward direction of data flow.
  \figurebox{J} In addition to interactively training the network, users can change the speed at which the visualization for cell steps advances.
  \figurebox{K} Just as in the network overview, users can modify training parameters.
  }
  \label{fig:celltraining}
\end{figure*}

In the following, we discuss the visualization design of~\enn{}.
We explain how we tackle the aforementioned visualization challenges \visualizationbox{Vx} and learning psychology goals \loadbox{CR/DU/MJ} while targeting the educational objectives \objectivebox{Ox} defined in~\autoref{sec:eduobj}.
We first describe the overall visualization concepts we implemented for~\enn.
Then, we elaborate on the different views of our environment in the upcoming subsections.
\\
\textbf{Scales.}
To show both an overview of the training process \objectivebox{O3} and give detailed insight into the computation that is performed within one recurrent cell \objectivebox{O2} \loadbox{DU}, we employ an overview first, zoom and filter, then details on demand visualization design, following Shneiderman's mantra \visualizationbox{V3}~\cite{shneiderman1996eyes}.
Therefore, \enn~consists of two main views, the network overview (\autoref{fig:networkview}), which displays the training progress on the network scale, and the LSTM cell view (\autoref{fig:celltraining}), which allows for a detailed inspection of an LSTM cell.
This is in line with our goal of reducing complexity \loadbox{CR} by focusing on individual steps of the learning process rather than presenting everything at once.
\\
\textbf{Animation.}
Animation has shown to be effective in visualizing data relationships and algorithms \objectivebox{O1-3}~\cite{bartram1997perceptual, chevalier2016animations, beck2014state}.
Furthermore, animation has shown to be associated with fun and excitement~\cite{robertson2008effectiveness}, which is in line with our goal of making learning more enjoyable \loadbox{MJ}.
Thus, to visually communicate how the network operates on sequential data, we use animation throughout our visualizations \visualizationbox{V2}.
\\
\textbf{Onboarding.}
Novel visualizations and interactive systems can be hard to understand~\cite{lee2015people}.
We designed \enn~in a way that allows exploration without running the risk of making irreversible errors or needing teacher supervision \visualizationbox{V4}.
However, instructional aids may be important to understand such complex content \loadbox{DU}~\cite{berthold2009instructional}.
Therefore, we use an onboarding process for our educational environment~\cite{kang2003new} (\cf{} \autoref{fig:onboarding}).
With this process, we aim to further reduce the cognitive load during learning compared to classical learning environments \loadbox{CR}~\cite{brachten2020ability}.
For example, the sequential nature of RNNs \objectivebox{O1} \visualizationbox{V2} and the data and tasks that RNNs can be used for \objectivebox{O4} are communicated in \enn.
\\
\textbf{Textual Explanations.}
In contrast to other learning environments, which show static textual explanations below the main visualization~\cite{smilkov2017direct,kahng2018gan}, we instead provide such additional information as \emph{details on demand} \visualizationbox{V3}~\cite{shneiderman1996eyes}.
This way, users can access more information for exactly the components they want to learn more about \loadbox{DU}, while not having to read a lot of text \loadbox{CR}.
Our interactively explorable dialog boxes, as shown in~\autoref{fig:explanation}, provide information about all important elements of the learning environment.
Such dialogs exist for all headings and are anchored through an \scalerel*{\includegraphics{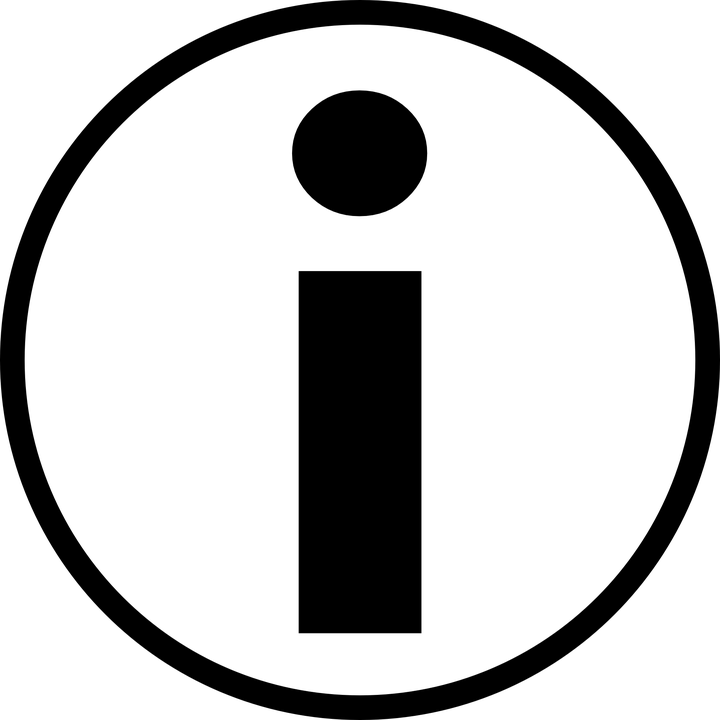}}{)} icon, and for all components of an LSTM cell, which is referred to in our onboarding process.

\subsection{Network Overview}

In the network overview, following the natural reading direction of western cultures, as well as related work on NN architecture visualization~\cite{drawconvnet,convnet_drawer,bauerle2019net2vis}, we arrange the network from left to right.
On the left, one can see the input type that is currently used to train the network \figurebox{A}.
Centered, we present an abstracted visualization of the network, where users can see how many layers the network contains \figurebox{B}.
On the right, a visualization of the prediction along with the prediction error shows how training progresses \figurebox{C}.
Below these visualizations, information about the training process, controls for the training process, and means to change training parameters are shown \figurebox{D-F}.
\\
\figurebox{A} \textbf{Input.}
To experiment with the network, users can select the input data that is used to train the network from a set of explanatory input types.
Data for an interactive and explainable visualization of NNs needs to both explain the network functionality \objectivebox{O1} and be easy to understand \visualizationbox{V1}.
Therefore, current educational visualizations use an abstract, two-dimensional distribution of points to train their networks on~\cite{smilkov2017direct, kahng2018gan}.
With \enn, we follow this approach of employing data that is as simple as possible \loadbox{CR}.
As RNNs are focused on sequential data, we decided to use periodical mathematical functions and simple text snippets, which map nicely to the sequential nature of RNNs \visualizationbox{V2}.
The functions that can be used as training data in \enn are a sinusoidal function, a sawtooth function, an oscillating function, and a composite sinusoidal function and vary in their periodicity.
To demonstrate the sequential and dynamic nature of these input functions, we animated those that are in use so that they seem to flow while being input to the network \loadbox{MJ}.
\\
In addition to abstract function continuation, we also provide text-based data to train the network on \loadbox{DU}.
To allow for interactive training, we employ rather simplistic text samples.
These include a recurring character sequence (\textit{ababab...}) and the well-known text \textit{lorem ipsum}.
Here, we employ a similar design language as with function data, to show that most ideas can be transferred across tasks.
By incorporating this text learning scenario, users of \enn~get to learn and inspect not only abstract problems, but can also experience more realistic scenarios \objectivebox{O4} \loadbox{MJ}.
\\
\figurebox{B} \textbf{Network.}
In the network visualization, we want to communicate the recurrent nature of our network \objectivebox{O1}, but at the same time, show all layers.
Thus, instead of the more frequently used unrolling of RNN layers~\cite{understand_lstm}, we add a loop to the layer glyph to symbolize this recurrence.
This symbolizes the feedback loop of information output at $t$ back to the input of a cell at $t+1$ \visualizationbox{V2}, which enables BPTT.
\begin{figure}[b]
    \centering
    \setlength{\fboxsep}{0pt}\fbox{\includegraphics[height=1.9cm]{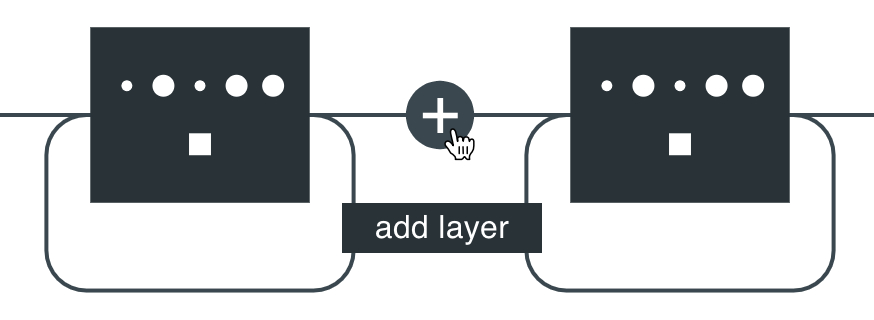}}
    \hfill
    \setlength{\fboxsep}{0pt}\fbox{\includegraphics[height=1.9cm]{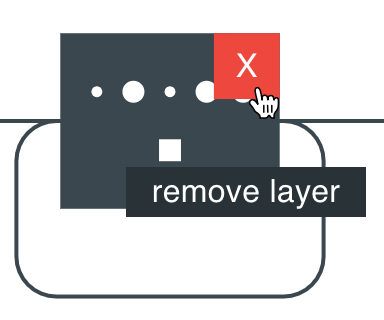}}
    \caption{\label{fig:add_remove}
      left: Adding a layer between two existing layers.
      right: Removing a layer from the network.
    }
\end{figure}
\\
Our network visualization is animated as data flows through its layers \objectivebox{O3} \loadbox{MJ}.
For the prediction step, dashed lines flow in the forward direction to symbolize forward data processing.
For the backpropagation step, they flow backwards to resemble the backpropagation of the error \visualizationbox{V2}.
Dashed lines are moving from input to output during the prediction phase, and from output to the first network layer during training, because backpropagation is not applied to the input domain.
\\
Users can also investigate how the training progresses differently depending on the number of recurrent layers in the network \objectivebox{O3}.
Therefore, layers can be added or removed from the network to be trained, as shown in~\autoref{fig:add_remove} \loadbox{DU}.
As with most explorable components, we explain the implications of this in our introduction, and users can click the \scalerel*{\includegraphics{information_icon}}{)} next to the network heading.
\\
\figurebox{C} \textbf{Predictions.}
Commencing the top row of visualizations is the data plot, where we visualize an input sample and the prediction of the network along with its ground truth \objectivebox{O3}.
Here, multiple data points that are processed by the network one after the other are used to inform a prediction, which is visualized by sliding a gray box over the input data that is currently processed \visualizationbox{V2}.
Additionally, the prediction values slowly build up with animations to clarify that this prediction is building up sequentially \loadbox{MJ}.
We then use vertical lines in the function plots, which slowly emerge between the prediction and the target value.
This vertical line encoding is in analogy with the way we calculate errors, namely, by looking at the prediction values and calculating the difference to the ground truth \loadbox{DU}.
The error calculation is embedded temporally between the inference (forward network animation) and backpropagation (backward network animation) phases of the training process \visualizationbox{V2}.
Altogether, through this animated component, while not being interactive itself, users can inspect the results of modifications that have been made in other places \objectivebox{O3}.
\\
\figurebox{D} \textbf{Process.}
According to the typical NN training setup, we divide the training process into three distinct steps: inference (forward), validation (error calculation), and backpropagation (backward).
The explanation pane in the lower left of the network overview (see~\autoref{fig:networkview}) displays which step is currently executed and provides an explanation of what happens in each of these steps \loadbox{DU}.
Through this, the user can learn more about the training dynamics of the network \objectivebox{O3}.
As described previously, animations in other components complement this dynamic nature of the training process \visualizationbox{V2}.
\\
\figurebox{E} \textbf{Controls.}
In the network overview, the network is trained by means of epochs to \emph{first} provide an \emph{overview}~\cite{shneiderman1996eyes} of the training process \visualizationbox{V3} \loadbox{CR}.
To experiment \objectivebox{O3}, users can interact with the control area in the bottom center of our environment \loadbox{MJ}.
In addition to automatically advancing epochs, which can be controlled with the \scalerel*{\includegraphics{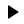}}{)} and \scalerel*{\includegraphics{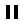}}{)} buttons, users can also trigger network training for a single epoch, by pressing the \scalerel*{\includegraphics{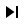}}{)} button \loadbox{DU}.
The training process can always be reset using the \scalerel*{\includegraphics{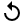}}{)} button.
A back button to go to a previous epoch is not included in \enn{} as this would require saving multiple previous states of the network parameters, which would require significant browser memory.
Therefore, and as individual epochs normally do not change the network behavior completely, going back one training epoch during training is not a common operation during neural network training, so we think users will not miss such functionality.
\\
\figurebox{F} \textbf{Training Parameters.}
To communicate the training setup of an RNN \objectivebox{O3}, a trade-off between completeness and simplicity must be made \visualizationbox{V1}.
Thus, we let the user freely choose some training parameters, but employ restrictions for others \loadbox{CR}.
As mentioned, users can add or remove individual network layers and use different preset training inputs.
In addition, they can change the learning rate, batch size, and noise \loadbox{DU}.
The learning rate and batch size allow for exploration of different training settings \objectivebox{O3}.
Noise can be added to make the training data more realistic, resembling real-world scenarios of imperfect measurements \objectivebox{O4}.
Parameter changes can be made through sliders, which are positioned on the bottom right.
To provide an intuition about the influence of these parameters, we include pretrained models that are loaded during the onboarding steps which explain each individual parameter \visualizationbox{V4}.
Other parameters, such as units per layer or optimization strategies, cannot be changed in our implementation.
This trade-off between freedom of exploration and simplicity proved to be effective in educating users about the influence of different training parameters and keeping their cognitive load low \visualizationbox{V1}.
\\
\\
Hierarchical aggregation can help simplify visualization designs \loadbox{CR}~\cite{elmqvist2010hierarchical}.
Thus, after getting an overview of the network, the user can inspect another hierarchy level in detail, namely individual LSTM cells \objectivebox{O2}~\cite{shneiderman1996eyes}.
When selecting one of the layers in the network overview a zooming transition onto one of the network layers gradually reveals the structure of an LSTM cell to support the user's mental image of looking into one of the layers \visualizationbox{V3} \loadbox{CR}.
With this multiscale approach, where users can navigate between views, orientation is important \loadbox{CR}.
Therefore, a color coding indicates the current level of detail.
This highlight color is blue for the network overview, whereas orange is used for the LSTM cell view.     
Orange and blue are complementary colors, which makes them easily distinguishable, and they can be differentiated by vision-impaired users~\cite{jenny2007color}.

\subsection{LSTM Cell View}
In the LSTM cell view, we show a detailed visualization of the selected cell on the left, embedded in small pictograms of neighboring cells \figurebox{G}.
On the right of this cell visualization, one can see the input, target, and prediction values of the network, where new points are added as they flow through the cell \figurebox{H}.
Below these visualizations, we show information about the training process, controls for the training process, and means to change training parameters, similar to the network overview \figurebox{I-K}.
\\
\figurebox{G} \textbf{Cell Architecture.}
To convey the functionality of one recurrent unit \objectivebox{O2}, we show all computational elements within a cell \loadbox{DU}.
Wherever information is combined, we show a \scalerel*{\includegraphics{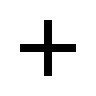}}{)} icon.
Icons for the input (\scalerel*{\includegraphics{input_black}}{)}), forget (\scalerel*{\includegraphics{del_black}}{)}), and output gates (\scalerel*{\includegraphics{output_black}}{)}) visualize the gating functionality of an LSTM cell.
While all gates that transform the data are depicted with circular icons, the cell state, which represents the saved state of the cell, is represented by a squared \scalerel*{\includegraphics{save_black}}{)} icon, illustrating the semantic difference between these components.
Each of these cell components can be selected to get a detailed explanation of its functionality, as shown in~\autoref{fig:explanation}, marking another level of detail in this visualization \visualizationbox{V3} \loadbox{CR}.
\\
\\
Data flow is visualized through connecting lines \objectivebox{O2} and step-by-step animations of the cell components \loadbox{MJ}.
Here, elements that process data in the currently visualized computation step are highlighted.
As in the network overview, connections moving data are symbolized with dashed lines.
Those lines flow forward during inference and backward during backpropagation.
This way, we communicate how the hidden state and output of these cells is computed and visualize how the data flows from one to the next operation or gate \visualizationbox{V2} \loadbox{DU}.
\\
The reverse data flow of BPTT occurs not only once within a cell to backpropagate to the previous layer, but multiple times, for all input time steps \objectivebox{O1}.
The connections within the cell also clarify that there are two recurrent cycles, one from the output of the cell back to the input, and one within the cell to update the cell state based on its state in the previous iteration \loadbox{DU}.
As a result, while other visualizations require~\textit{unrolling}, where time steps are visualized by displaying multiple cells in concatenation~\cite{understand_lstm}, we communicate recurrence through step-by-step animation.
This removes the ambiguity of stacked layers vs. unrolled cells, which was shown to hinder learners in our first experiments \visualizationbox{V1} \loadbox{CR}.
\\
\figurebox{H} \textbf{Data Plot.}
Right of the cell visualization, we show the input data, network prediction, and ground truth all in one graph.
In contrast to the network overview, where the network is directly connected to this output graph, the cell is disentangled from this visualization.
As the depicted cell typically receives data from previous cells and outputs data to subsequent cells, this visualization, where animation steps are synchronized but not visually connected on both the input and output side, better reflects the network architecture of RNNs \visualizationbox{V3} \loadbox{DU}.
Users can inspect this view during interactively controlled training to see how the network processes input data to make predictions sequentially and how it calculates the training loss in relation to the processing steps within a cell \objectivebox{O1\&2} \visualizationbox{V2}.
\\
\figurebox{I} \textbf{Training Process.}
The three steps of inference, validation, and backpropagation are just as relevant in the LSTM cell view as they are in the network overview \objectivebox{O3}.
As the training speed is lower in the LSTM cell view, users can skip part of the data processing and go directly to the processing step of interest \visualizationbox{V1\&2}.
For the forward pass, we add additional explanations for the different processing steps of \textit{receiving the layer input}, \textit{calculating the gate activations}, \textit{updating the cell state}, and \textit{outputting the activation value} \loadbox{DU}.
These explanations are highlighted in synchronization with the processing steps during the forward pass to the data flow in the cell visualization above \loadbox{MJ}, allowing users to draw links between the processing steps and the explanations they are interested in \visualizationbox{V2\&4} \loadbox{CR}.
\\
\figurebox{J} \textbf{Controls.}
In the LSTM cell view, processing is done by means of compute steps, showing a much more detailed processing pipeline than in the network overview \visualizationbox{V3}.
As in the network overview, the control area can be used to experiment with the training process \objectivebox{O3}.
The more fine-grained advancement of the visualization is also adopted by the degree to which the animation advances with the forward button, since it only executes the next compute step within a cell \loadbox{DU}.
In addition to what can be done in the network overview, the speed of the animations for data processing within this cell can be adjusted, so that users can explore the processing steps at their own pace \visualizationbox{V2}.
\\
We want to emphasize the buildup of state within a cell based on multiple input time steps.
Thus, we show how the network processes these inputs in great detail, whereas we made the animation of the backpropagation take less time than forward processing.
As \enn{} is not designed to represent accurate timings anyway, this is our way of visualizing cell processes in detail, while also preserving the ability to observe multiple epochs.
\\
\figurebox{K} \textbf{Training Parameters.}
Training parameters can be adjusted in the LSTM cell view just as in the network overview, giving the user even more control over the training process and room for experimentation \objectivebox{O3}.
\\
\\
To get back to the network overview, one can click anywhere outside of the LSTM cell in~\autoref{fig:celltraining} \figurebox{G} \visualizationbox{V3}.

\section{Technical Realization}
While the visualization design described above has been carefully crafted to meet the educational objectives described in~\autoref{sec:eduobj} and the visualization design challenges outlined in~\autoref{sec:vischallenges}, its technical realization needs to take into account the technical challenges identified in~\autoref{sec:techchallenges}.
In this section, we detail how we tackled these technical challenges.
\\
\textbf{\technicalbox{T1} Training Time.}
While an RNN for a complex application cannot be trained live in the browser, we simplify the problem in multiple ways.
By employing simplistic data sets, the model can converge after relatively few epochs.
Additionally, we limit the number of data points that are fed to the network per epoch.
Therefore, epochs are processed sufficiently fast for our interactive visualization approach.
We also limit the network size to at most seven layers, so that memory consumption and processing time are reduced.
In turn, users can see the training progress and get visible prediction improvements after only a few epochs, while one such epoch takes seconds to compute.
\\
\textbf{\technicalbox{T2} Training Steps.}
A key aspect of our approach is the decoupling of computation and visualization.
Through this decoupling, we are able to show the training steps in an observable manner and enable exploration at the user's own pace.
This helps users understand how the model processes input data and predicts new data points.
\\
\textbf{\technicalbox{T3} Deployment.}
To be able to make~\enn~publicly available for a large audience, we implemented it as an interactive browser application using HTML and JavaScript.
To train the RNN, we use TensorflowJS~\cite{smilkov2019tensorflow}, for animated visualizations of the trained network, we use P5.js~\cite{p5js}.
This way, we are able to provide an interactive, web application that visualizes the training dynamics of RNNs through animation, which is accessible at:~\url{https://mi-pages.informatik.uni-ulm.de/explornn/}.

\section{Limitations}\label{sec:limitations}
While \enn~provides a novel environment for learning about RNNs, there is still room for more advanced visualization designs that could be explored in the future.
Some of these limitations are explained in the following.
\\
\textbf{Explanations.}
\enn~offers a lot of experimentation that is complemented by textual explanations.
However, the number of textual explanations that fit into the context of such an educational system, which is designed to provide an overview of this complex topic, is insufficient to fully explain RNNs.
For specific questions that are not addressed by our interactive system, we refer to developer documentation and scientific papers.
\\
\textbf{Drill-Down.}
\enn~explains RNNs on both a network and a cell level.
Apart from seeing the data flow on these granularities and textually describing the components of a cell, visualizing the workings of these components could further benefit the learning experience.
However, these components are just mathematical functions to which neither the input nor output have a directly discernable meaning.
If we were to, \eg{} visualize the internals of a memory component, users could only see matrices of seemingly meaningless numbers flowing through these cells.
This would not add any benefit and might even result in confusion about such a visualization.
To explain these internal components, novel interpretability techniques might help.
Inventing and implementing those is beyond the scope of this work.
\\
\textbf{Component Change.}
To see the influence of individual components in a cell, changing or removing them could be an interesting addition to the workflow.
We did not implement this capability for two reasons.
First, adding such functionality goes deep into the working of individual cells, which would exceed the learning objective of getting an overview of RNNs and LSTM cells.
In turn, we assume that changing single components in individual cells is unlikely to have a measurable and interpretable effect on the overall learning outcome.
Second, we would have needed to implement our own DL library for this to be possible, as TensorflowJS has predefined LSTM layer implementations.
\\
\textbf{Degrees of Freedom.}
While users can change some hyperparameters and network settings in our environment, we deliberately do not expose all possible settings to our users.
The goal of this limited exploration setting is that users can get an overview of important manipulations to be made, while at the same time not overwhelming our target audience.
As for limited explanations, we refer to developer documentation and scientific papers for users that want to explore these details.
\\
\textbf{Layer Types.}
In our implementation, we focused on conveying LSTM cells.
However, there are numerous other cell architectures for RNNs.
Although we don't think this limited focus hinders learners with understanding RNNs on a high level, it would, nonetheless, be helpful for users specifically interested in certain cell types to include these in \enn.

\section{User Study}
To evaluate the effectiveness of our approach, we conducted a user study with 37 participants (30 male, 7 female) aged between 21 and 32.
Participants were recruited from a DL course at our local university.
Our study was a lecture at the end of the course, after students had already learned about feed-forward NNs.
Participants were randomly assigned to one of two groups.
The \enn~group received the interactive application, and the text group was presented a text-based learning environment.

To look at learning outcome in detail, our evaluation was divided into the first three distinct, hierarchical cognitive learning goals according to Bloom's taxonomy~\cite{bloom1956taxonomy}, namely recall, comprehension, and transfer.
We expect higher learning outcomes for the \enn~group compared to the text group at all three levels.
For a closer look at the cognitive processes involved in learning \loadbox{CR}, we also collected data for the three types of cognitive load~\cite{sweller2011cognitive}.
Intrinsic cognitive load (ICL) results from the natural complexity that underlies the learning content.
Since the difficulty does not differ, there should be no difference between the two groups.
Extraneous cognitive load (ECL) is caused by inadequate instruction or presentation of information.
Due to the step-by-step presentation of information and the direct connection of textual information and explanatory figures in the \enn~group, we expect lower ECL for the~\enn group compared to the text group.
Lastly, germane cognitive load (GCL) represents the invested learning-related load.
GCL is connected to the processes that are needed to construct and automate mental representations~\cite{sweller2011cognitive}.
Following the reduced ECL in the \enn~group, learners should have more free cognitive capacity in working memory to invest in learning-related GCL.

\subsection{Hypotheses}
Based on the described theory, we hypothesize the following.
We expect a higher learning outcome, differentiated by recall \hypothesisbox{H1}, comprehension \hypothesisbox{H2} and transfer \hypothesisbox{H3} in the \enn~group than in the text group.
Furthermore, we expect no differences between the groups for ICL \hypothesisbox{H4}.
We expect a lower ECL in the \enn~group than in the text group \hypothesisbox{H5}.
For the GCL, we expect it to be higher in the \enn~group compared to the text group \hypothesisbox{H6}. 

\subsection{Method}
Our study was split into different steps, which we explain in the order they were presented to the participants.
\\
\textbf{Prior knowledge.}
Prior knowledge was measured with seven open-ended questions on NNs and DL techniques (e.g., \emph{Name two activation functions used in deep learning.}).
The questions were developed by a domain expert.
All answers were rated by a domain expert, following a predefined solution to ensure objectivity.
A total of one point could be scored for each question, with partial points of .5.
The maximum score for the prior knowledge test was seven. 
\\
\textbf{Motivation (MSLQ).}
To assess motivation, the MSLQ~\cite{pintrich1991manual} subscale for motivation was used.
The MSLQ is a self-report questionnaire designed for an academic setting.
Motivation was measured with twelve items (e.g., \emph{I'm confident I can do an excellent job on the tests in this study.}).
Learners were instructed to respond as accurately as possible, reflecting their attitudes and behaviors towards the learning module.
Responses were given on a 7-point Likert scale ranging from 1 \emph{strongly disagree} to 7 \emph{strongly agree}.
Cronbach’s Alpha was computed for the internal consistency of the measures~\cite{cronbach1951coefficient}, the reliability was $\alpha = .95$. 
\\
\textbf{Learning material.}
The learning material was presented either as a text with illustrating figures, formulas, and graphs (see our supplementary material) or through \enn (see website).
For both conditions, the information was the same.
The only difference was the presentation medium and the lack of interactivity in text-based learning.
\\
\textbf{Learning outcome.}
To assess learning outcome, a domain expert developed a post-test with 11 open questions on the content of the learning session.
To better understand cognitive processes, the posttest was differentiated by the first three levels of Bloom’s taxonomy \loadbox{DU}~\cite{bloom1956taxonomy}.
Recall was measured with four questions (e.g., \emph{Name the backpropagation algorithm that is used for RNNs.}).
Comprehension was also measured with four questions (e.g., \emph{Explain the meaning of this formula: $c_t = filtered\_input + filtered\_state$.}).
The main purpose of these questions was to test how well people could explain and discuss the learning content.
Transfer was measured with three questions (e.g., \emph{Assuming you have a poem continuation network and training data with poems from the internet. If your network now makes a prediction, how do you determine if it is correct, to calculate the loss?}).
These questions were designed to test the ability of learners to draw inferences from the learning content and apply it to new contexts.
Similarly to the prior knowledge test, each question was rated by a domain expert, following a predefined solution to ensure objectivity.
A total of one point could be scored for each question, with partial points of .5.
The maximum score for recall and comprehension was four each, and for transfer, it was three, so the total maximum score for the post-test was eleven. 
We did an ANOVA on the learning outcome to test for statistical significance.
\\
\textbf{Cognitive load.}
To measure cognitive load \loadbox{CR}, the differentiated cognitive load questionnaire was used~\cite{klepsch2017development}.
It contains two items for ICL, three items for ECL, and three items for GCL, all measured as self-reports on a 7-point Likert scale from 1 \emph{strongly disagree} to 7 \emph{strongly agree}.
To measure internal consistency, the Cronbach Alpha was calculated~\cite{cronbach1951coefficient}.
Reliability was $\alpha=.66$ for ICL, $\alpha=.81$ for ECL, and $\alpha=.77$ for GCL.
As for learning outcome, we tested for significance with an ANOVA.
\\
\textbf{System usability.}
To quantitatively measure the system usability, the System Usability Scale (SUS) was used \cite{brooke1996sus}.
This scale is a self-report measurement consisting of 10 items related to the usability of \enn (e.g., \emph{I found the system very cumbersome to use.}).
Responses to the items were given on a 7-point Likert scale ranging from 1 \emph{strongly disagree} to 7 \emph{strongly agree}.
The internal consistency of this scale was $\alpha=.74$.
\\
\textbf{Qualitative questions.}
For an impression of the quality of the learning material, further questions were implemented.
Three open-ended questions were related to likeability (\emph{What about the learning experience did you like especially, what did you not like?}), missing functionality (\emph{Was there something you would have liked to do but could not?}), and additional comments (\emph{Other remarks.}) \loadbox{MJ}.
For liking (\emph{I would like to use this learning material.}) and recommendation (\emph{I would recommend this learning material to my friends.}) of the material, two items could be rated on a 5-point Likert scale from \emph{very unlikely} to \emph{very likely}.
Content (\emph{How was the quality of the content?}) and design (\emph{How was the design of the learning experience?}) could be rated with 0 – 5 stars.  

\subsection{Results}
In the following, we present the results of the user study.
\\
\textbf{Descriptive data.}
The analysis of the descriptive statistics showed that subjects of the text group and the \enn~group did not differ in most of the variables.
T-tests (variances were equal for all variables) with respect to age ($p=.33$), gender (\enn~ group 21\% females, text group 16.67\% females) ($p=.74$), MSLQ ($p=.11$), self-efficacy (MSLQ) ($p=.16$) and duration ($p=.79$) revealed no significant differences.
Motivation (MSLQ) showed a significant t-test ($p=.02$), indicating that learners in the text group had a significantly higher score.
Descriptive data for all variables per condition can be seen in~\autoref{tab:descriptive}.
\\
To analyze whether prior knowledge and MSLQ should be used as covariates, we conducted a correlation analysis with learning outcomes and cognitive load.
Significant correlations could be found for GCL with the MSLQ ($r=.37$, $p=.024$) and for the recall of the post-test with the MSLQ ($r=.44$, $p=.006$).
Therefore, they were included as a covariate in the following calculations concerning GCL and recall.
No other significant correlations for the potential covariates could be found.
\\
\textbf{Learning outcome.}
Against our hypotheses, we found a significant difference regarding recall ($F(1, 34)=3.91$, $p=.028$, $\eta_p^2=.103$) in favor of the text group but not for comprehension ($F<1$, n.s.) or transfer ($F<1$, n.s.).
\\
\textbf{Cognitive load.}
Contrary to our expectations, we found a significant difference between text and \enn~group for ICL ($F(1, 34)=3.85$, $p=.029$, $\eta_p^2=.099$).
ECL showed the hypothesized effect: The \enn~group showed a significant lower score than the text group ($F(1, 34)=4.33$, $p=.023$, $\eta_p^2=.113$).
Against our hypothesis, GCL was not significantly higher in the \enn~group than in the text group ($F<1$, n.s.).
\\
\textbf{System usability.}
The SUS questionnaire indicates an \emph{excellent} usability~($M=84.47, SD=9.45$)\cite{bangor2009determining}.
Participants also rated our approach as significantly more likable ($F(1, 30)=10.52$, $p=.003$, $\eta_p^2=.260$), more recommendable ($F(1, 30)=11.75$, $p=.002$, $\eta_p^2=.281$), and better designed ($F(1, 30)=20.711$, $p<.001$, $\eta_p^2=.408$) compared to the learning text.
\\
\textbf{Qualitative questions.}
We also got some qualitative feedback in our free-form fields.
Participants liked our introduction, which apparently made it easy for them to get started with \enn{} \emph{the tutorial was nice and the platform was easy to use}.
They also mentioned that the graphical support of these textual explanations was helpful for them to form a mental image of the setting: \emph{the mental bridge the graphical presentation helped build was helpful in memorizing and understanding}.
Some participants said that they did not remember specific names, as it was not important during the usage of \enn{}: \emph{I later did not remember the name of the algorithm that was used, since it was not important during the usage of the tool}.
Some participants asked for something similar for other types of networks, e.g. \emph{I would like to have similar resources to cover other topics from the basics such as MLPs up to advanced topics and more complicated kinds of networks}.
As described in \autoref{sec:limitations}, we only support a limited set of interactions, which some participants commented on, e.g. \emph{[I missed] changing the activation function of the LSTM gates}.

\begin{table}[t]
  \centering
  \caption{
    Means and standard deviations for our study results separated by groups for all variables.
    Numbers annotated with * indicate a significant difference between the two conditions.
  }
  \label{tab:descriptive}
  \begin{tabular}{l|ll|ll|l}
  	& \multicolumn{2}{l|}{\textbf{Text}} & \multicolumn{2}{l|}{\textbf{\enn}} \\
  	\textbf{Variable} & \textbf{M} & \textbf{SD} & \textbf{M} & \textbf{SD}\\
  	\hline
  	\hline
  	Duratiton (min): & 43.6 & 26.78 & 40.87 & 35.67\\
  	Age (years): & 24.94 & 2.96 & 24.05 & 2.51\\
  	Pre-T: (\%) & 83.14 & 13.86 & 80.0 & 14.57\\
  	Post-T: (\%) & 68.48 & 20.46 & 65.47 & 14.41\\
  	Post-T recall*: (\%) & 79.75 & 26.30 & 63.25 & 29.64\\
  	Post-T comp.: (\%) & 65.00 & 51.54 & 65.50 & 37.26\\
  	Post-T trans.: (\%) & 58.00 & 69.54 & 68.33 & 41.46\\
  	ICL*: (\%) & 61.51 & 18.51 & 48.82 & 20.51\\
  	ECL*: (\%) & 48.14 & 23.32 & 37.09 & 15.52\\
  	GCL: (\%) & 77.51 & 13.94 & 75.94 & 17.13\\
  	MSLQ*: (Max: 7) & 5.17 & 0.95 &  4.56 & 1.30\\
  	SUS: (Max: 100) & - & - & 84.47 & 9.45\\
  	Qualit. Like*: (Max: 5) & 3.27 & 1.03 & 4.35 & 0.86\\
  	Qualit. Recomm.*: (Max: 5) & 3.00 & 1.20 & 4.24 & 0.83\\
  	Qualit. Content: (Max: 5) & 4.00 & 0.76 & 4.35 & 0.61\\
  	Qualit. Design*: (Max: 5) & 3.20 & 0.94 & 4.47 & 0.62\\
  \end{tabular}
\end{table}

\subsection{Discussion}
In the following, we will refer back to the hypotheses we had before conducting the study and discuss the study outcome.
\\
\textbf{Learning outcome.}
We looked at both recall and understanding regarding the learning outcome.
In contrast to \hypothesisbox{H1}, we found that learners in the text group showed significantly better results for recall.
While we found no significant differences between the groups regarding comprehension \hypothesisbox{H2} and transfer \hypothesisbox{H3} the results are interesting nonetheless.
Although not significant, the descriptive statistics indicate that the score for transfer is about 10\% higher for \enn{} compared to text \loadbox{DU}.
This could be a first indication that learning environments such as \enn{} can help learners build a deeper understanding of the subject compared to learning with classic text.
However, significant results and further research are needed to support this statement.
Compared to recall, these results may indicate that while learners are better at learning terms by heart (surface learning) when they learn with text than with~\enn.
\\
A possible explanation for the better recall performance in the text group could be that learners have more experience with text-reading strategies~\cite{kolic2011role}.
This might help with the complex terms explained here, as learners might find it easier to find information that was previously presented~\cite{kurschner2006konstruktion}.
Thus, designing ways to easily retrieve previously presented information could be an interesting direction of future research for such interactive explorables.
Another possibility is that learning with an interactive environment, which is affirmative and provides information step by step, might infuse the illusion of knowing~\cite{glenberg1982illusion}.
Learners may think that after a few experiments in \enn~they have acquired enough knowledge, while there is still much more to explore and learn.
In the text group, it is immediately clear to the learner when the text is finished.
On the contrary, \enn can require user initiative for information acquisition.
As the learning experience was self-controlled, participants could decide for themselves when to go from the learning content to the post-test.
Referring to the illusion of knowing, learners might have felt too competent as they experienced this more guided experience.
However, even though learners may be able to transfer what they have learned to other application areas, they may be missing important basic terminology that was presented in the learning material to reflect their knowledge gain in a classical learning test. 
\\
\textbf{Cognitive load.}
Against \hypothesisbox{H4}, the \enn~group showed significantly lower ICL than the text group. 
The perceived difficulty of the learning material is 12.71\% lower for \enn~even though the text content was identical in both conditions.
This suggests that \enn~makes the learning material appear easier \loadbox{CR}.
The reason for this could be that the content is presented step by step in the tutorial of \enn, instead of all at once as in the text condition \cite{brachten2020ability}.
\\
The results regarding \hypothesisbox{H5}, are consistent with our assumptions.
With a large effect size, the \enn~group showed lower ECL than the text group \loadbox{CR}.
Therefore, \enn reduced the extraneous cognitive load compared to the text content, although the content was the same in both conditions and there were no unnecessary figures or information in the text.
Combined with lower ICL, more cognitive capacity remains for GCL, which is important for learning.
\\
For GCL, we did not find the significant difference we hypothesized in \hypothesisbox{H6}.
Although ICL and ECL indicate that more cognitive capacity should be free in the \enn~condition, participants did not invest that cognitive capacity into GCL.
This could be because there might already be high investment in GCL in both groups.
Another explanation could be that since participants in the text group perceived the learning content as more difficult, they might have invested more GCL to compensate for said difficulty.
\\
\textbf{System usability.}
The results of the SUS questionnaire indicate that our system is easy to use.
This supports our proposed visualization and interaction design and shows that our design choice of creating an interactive environment as a learning experience on RNNs matches our target audience well.
Additionally, as participants rated our approach as significantly more likeable, recommendable, and better designed, users are likely to experience more joy, and be more motivated when learning \loadbox{MJ}.
In combination with the reduced cognitive load \enn{} inflicts on users, we hope that this could result in a larger number of users willing to spend their time learning and more time spent learning per user.
In turn, we think that this might outweigh the advantage in some areas of learning outcome with the more familiar text-based learning environment.
Further longitudinal studies on NN learning systems need to be done to investigate this in more detail.
\\
\textbf{Qualitataive feedback.}
The open feedback forms also provided interesting insights.
In general, participants seemed to like \enn{} as a learning experience and even asked for similar interfaces in other contexts.
Furthermore, the amount of information and our onboarding process seemed to make \enn{} easily usable.
Most of the criticism was related to limitations regarding the freedom of interaction, which we deliberately implemented to provide an overview rather than in-depth technical details.
Future work might reveal how both an overview and full depth could be combined in NN learning environments.
We also learned why recall might be better in the text condition.
As participants mentioned, they did not feel they needed to memorize specific terms to be able to use RNNs.
This seems natural, as when programming or using RNNs in the wild, remembering specific terms is also not essential as they can be searched for.
On the other hand, transfer tends to be much more important when tasked with solving real problems.

For learning material where transfer is important~\loadbox{DU}, as in recurrent networks, our descriptive results suggest that interactive visualizations such as \enn~might be helpful.
Additionally, the lower cognitive load and higher perceived likeability of our interactive environment might result in more learners spending more time with \enn.
Although we extensively evaluated \enn~in this study, it remains to be seen whether our insights are transferable to other learning environments.
If so, the development of future explorables could be much better informed, indicating what is important, what could be discarded, and what needs to be improved on.
While this study provides first insights into the effectiveness of such educational NN exploration environments, we hope that similar evaluations of other applications can broaden our insights.

\section{Conclusions and Future Work}

This paper presents the first interactive learning environment specifically designed for RNNs.
We propose a new visualization approach for inspecting RNNs where different levels of granularity are employed.
To inform our visualization design, we first introduced educational objectives for this setting.
Based on these objectives, we identified design challenges, which we tackle in the proposed learning environment.
We hope that this process can be helpful for the development of future interactive learning environments.

Subsequently, we tested the learning outcome, cognitive load, and usability of this learning environment in an empirical study.
Our study is the first quantitative evaluation of an interactive NN learning interface and, as such, provides helpful insights and directions for future work.
The results of the user study indicate that, while the raw learning outcome was not improved compared to conventional methods, \enn{} makes learning easier and more fun since cognitive load was significantly reduced by \enn{} while subjective likeability was significantly improved.
Based on these findings, we propose to specifically design interactive NN learning environments so that cognitive load is reduced.
For broadly accessible education, \enn~can be used in any modern browser at~\url{https://mi-pages.informatik.uni-ulm.de/explornn/}.

As mentioned, more user studies for similar educational explorables could further advance the field and better inform future visualization designs.
One such possible research direction is the suspected advantage of the text condition for going back to previously presented information.
Here, eye-tracking studies and novel interaction designs might provide new insights.
Additionally, it would be interesting to investigate whether such systems indeed lead to more voluntary learning and how that affects learning outcome.

\section*{Statements and Declarations}
This work was funded by the Carl-Zeiss Scholarship for Ph.D.~students.
\\
The datasets generated during and/or analysed during the current study are available from the corresponding author on reasonable request.

\ifCLASSOPTIONcaptionsoff
  \newpage
\fi

\IEEEtriggeratref{58}


\bibliographystyle{IEEEtran}
\bibliography{exploRNN}

\newpage

\end{document}